# QuantFlow: A Federated Mamba-Based Post-Transformer Model for Time-Series Forecasting

**Shah Nawaz Haider** [1], **Steve Austin** [1], **Arnab Barua** [2], **Hadaate Ullah** [3], **Sarowar Morshed Shawon** [2*]

[1] Department of Computer Science and Engineering, University of Science and Technology Chittagong
[2] Department of Electrical and Electronic Engineering, University of Science and Technology Chittagong
[3] Faculty of Science, Engineering and Technology, University of Science and Technology Chittagong
nawazhaider60@gmail.com, steve_austin_@outlook.com, arnab.me@outlook.com, hadaate@ustc.ac.bd, shawon@ustc.ac.bd

**Abstract**

Time-series forecasting under heterogeneous, privacy-sensitive, and large-scale environments remains challenging due to the computational limitations of Transformer-based architectures and the centralized nature of most existing foundation models. This paper introduces QuantFlow, a unified federated forecasting framework which incorporates inverted sequence embedding, bidirectional Mamba state-space modeling, TSMixup augmentation, probabilistic quantile regression, and sample-weighted federated optimization, allowing for scalable and uncertainty-aware, yet privacy-preserving forecasting. The unified architecture replaces quadratic-complexity attention by linear-complexity state-space modeling for capturing long-range temporal dependencies, with preserved inter-variable dependencies. Our experiments on 13 (8 for pretraining, centralized and federated evaluations and 5 for zero-shot predictions) diversified real-world datasets, covering multiple domains such as energy, transportation, meteorology, finance, environmental monitoring and epidemiology, demonstrate that our framework performs accurately under centralized training, with $R^2>0.90$ achieved for all 8 datasets used in centralized learning. In a federated setting with 20 non-IID clients and 3 communication rounds, QuantFlow is also effective while maintaining competitive accuracy. Comparison across forecasting horizon of up to 720 steps also illustrate QuantFlow exhibiting comparable, or even better performance than standard models like Informer, FEDformer, TimesNet, TimeMixer, iTransformer and N-HiTS. Zero-shot experiments reveal strong cross-domain generalization, with an average $R^2$ of 0.760 on Solar-Energy and 0.752 on Bitcoin. Ablation study confirms that both inverted sequence embedding, and bidirectional Mamba layer contribute positively to the forecasting performance. Unlike existing approaches, QuantFlow unifies inverted sequence embedding and bidirectional Mamba within a federated framework, enabling linear-complexity, uncertainty-aware forecasting with state-of-the-art accuracy and strong zero-shot transferability. These findings position QuantFlow as a scalable and efficient foundation framework for high-accuracy, uncertainty-aware, and privacy-protected time-series forecasting applications.

**Code:** https://github.com/nawaz0x1/QuantFlow

## 1 Introduction

Artificial intelligence is reinventing forecasting by learning the intricate temporal patterns in data, artificial-intelligence-driven time-series forecasting tools have the potential to be applied in virtually every industry. However, more recent deep learning (DL) technologies have enabled improved forecasting of time series data that is so extensive, and yet often so volatile, as to have been intractable using classical statistical approaches. Addressing these, foundation model (FM) work by having reusable representations that are learned across heterogeneous large datasets instead of rebuilding task-specific models from scratch. This principle benefits time-series forecasting since organizations repeatedly predict various figures such as stock prices, electrical demand, traffic occupancy, disease prevalence, equipment status, or weather data. This approach could also reduce manual feature engineering and offer a consistent forecasting interface. Even so, the existing time series FM faces three challenges which hinders its practical use.

As per computational scalability, most time-series FM are based on Transformer attention, and both memory and computation cost increase with data dimension and context size.

When it comes to data locality, the diverse datasets used to train most foundations models can be distributed among organizations, who may not be able to share raw observations due to regulations, security, ownership or privacy.

Federated learning (FL) resolves the data locality issue through local training and global model aggregation rather than sharing raw data, while it does not address the computational bottleneck since it relies on the same communication of updates as in centralized settings (McMahan et al. 2017). Selective state-space models (SSM) like Mamba mitigate this bottleneck by learning an input-dependent SSM which linearly scales with sequence length and avoids recurrent computations (Gu and Dao 2024).

The model QuantFlow presented here is a time-series FM that incorporates state-space sequence processing with probabilistic forecasting and distributed training. We use an inverted embedding strategy so that every variable is represented across the entire observed sequence, 6 bidirectional Mamba decoder layers to model relationships among variables, and a quantile head which allows us to report both median prediction and prediction intervals. TSMixup enhances temporal data diversity, whereas we use a 20-client non-IID FL setup to evaluate the accuracy of our model on real data across domains, using limited communication rounds between client-server pairs without transmitting any raw observations. The study provides an integrated method for scalable and privacy-preserving forecasting across domains while acknowledging some of its limitations, like performance on irregularly sampled data and on long horizon prediction.

## 2 Related Work

This section covers the history of time series forecasting from the traditional methods up to recent trends, such as DL and FM, FL and SSM. We mainly touch on method that are more relevant to scalable probabilistic and distributed forecasting.

Traditional autoregressive, exponential smoothing and Theta models remain popular due to their interpretability and ease of understanding (De Gooijer and Hyndman 2006; Hyndman and Athanasopoulos 2018). However, training and managing models becomes complicated when dealing with large numbers of related series that need to be estimated and tuned individually. Deep forecasting allows for learning of representations directly. For probabilistic autoregressive time-series forecasting of multiple related series, DeepAR has been proposed (Salinas et al. 2020), and residual feed-forward architectures can achieve strong performance without recursion, such as in N-BEATS and NHITS (Oreshkin et al. 2020; Challu et al. 2023). Recent surveys continue to highlight the importance of evaluation protocols, dataset size and distribution shift (Benidis et al. 2022).

Long-term and cross-variable dependencies in time-series forecasting can be learned by Transformer models. The Temporal Fusion Transformer captures multiple input features and enables multi-horizon forecasting, the Informer model decreases attention costs and computation for long input sequences, and the model uses tokens consisting of temporal patches (Lim et al. 2021; Zhou et al. 2021). By reversing the standard input tokenization iTransformer enables direct modelling of cross-variable interactions (Liu et al. 2024).

Two dimensional and multiscale representations can also be achieved without Transformer attention, for instance using TimeMixer and TimesNet (Wang et al. 2024; Wu et al. 2023). These are foundations models because they leverage pretraining on large, heterogeneous corpora and enable transfer learning across datasets. Chronos treats time-series forecasting as token generation by a decoder-only model, and multiple FM offer universal interfaces that reduce reliance on retraining (Ansari et al. 2024; Das et al. 2024; Woo et al. 2024). Most of these approaches require centralized training, and many can struggle with high-dimensional input data.

The standard Federated Averaging algorithm operates by iteratively broadcasting a global model to the clients, having them run local training on their data and sending their model updates to the server for aggregation according to local sample counts (McMahan et al. 2017). Although raw data stay at the clients, the non-IID distributions at each client site (which might have different data scales, seasons, patterns or event frequencies) can result in clients drifting apart. SSM offer a complementary approach to efficient sequence learning. The work in Rangapuram et al. (2018) used Deep SSM for probabilistic time-series forecasting by combining them with dynamical systems. In Gu and Dao (2024) the Mamba model is introduced, which is based on input-dependent state-space transitions and has a linear complexity with respect to input length.

QuantFlow utilizes forward and reverse Mamba sequences, each within the observation window, so both directions are available at the point of prediction.

## 3 Methodology

This section presents the design and implementation of QuantFlow, from preprocessing and data augmentation to probabilistic prediction and federated training. The complete workflow and its principal components are described in the following subsections.

### 3.1 Framework Overview

QuantFlow is an end-to-end forecasting pipeline, which is comprised of data processing, augmentation, probabilistic representation learning and federated optimization. The original time series observations are preprocessed with chronological sorting, missing value imputation, and calendar features added to capture cyclic patterns. Then periodic features are represented as cyclical encoding, and variables are standardized before dividing them into fixed length input windows. TSMixup augments more diversity through weighted linear interpolation of standardized sequences, maintaining original sequential dependency.

For every augmented window, it first applies inverse embedding, and represents variables as tokens representing the entire observation history. The transformed sequence is then fed into six bidirectional Mamba decoder blocks, where long-range dependencies, cross-variable correlations can be captured bidirectionally. A quantile regression head generates the median prediction along with several prediction intervals to represent the uncertainties. Lastly, the centrally trained and configured model is deployed on 20 non-IID clients and updated for 3 rounds using sample-weighted federated aggregation without sharing the raw data of clients. The overview of the whole process is illustrated in Figure 1.

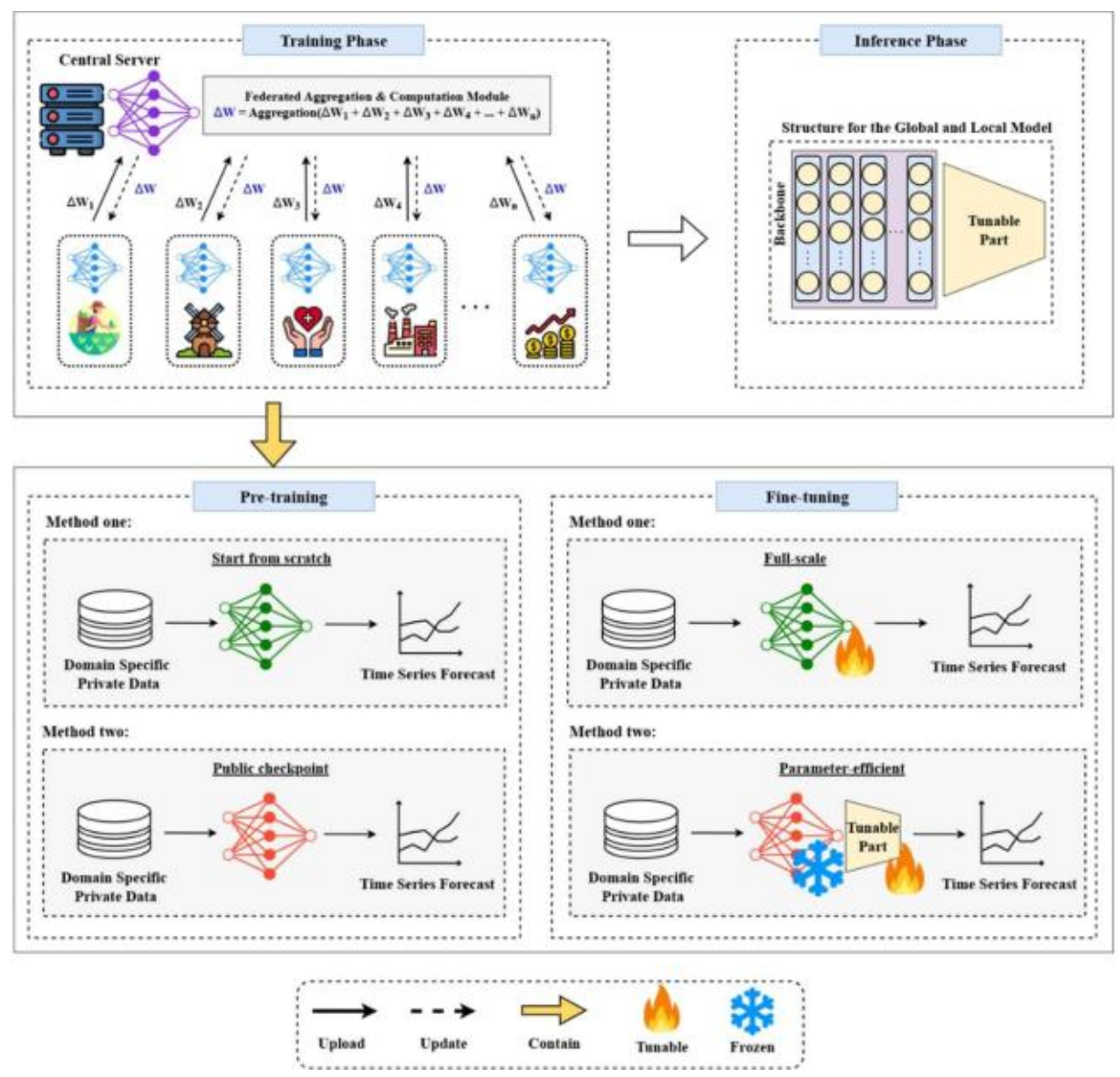

Figure 1: Federated training workflow, model structure, and strategies for pre-training and fine-tuning.

## 3.2 Data and TSMixup

We evaluate the framework on datasets from six application areas: Electricity (15 and 60-minute, 321 clients), Weather (21 meteorological variables, 60-minute resolution), Electricity Transformer (1-hour (h1, h2); 15 minutes (m1, m2) for oil temperature and power load, respectively) broken into ETTm1, ETTm2, ETTh1 and ETTh2, San Francisco Traffic (hourly freeway occupancy) and Global Temperature; which were used in pretraining, centralized evaluation and FL evaluation.

Cryptocurrencies (Bitcoin, Ethereum, Solana (Open, High, Low, Close, Volume)), Influenza-Like Illness (ILI) and Solar-Energy (10 minutes, 137 PV, Alabama State) datasets were used for testing zero-shot performance.

TSMixup generates a normalized sequence by sampling from an existing set of k existing sequences, $i_1, ..., i_k$ with weights 1, ..., k and sum(1, ..., k) = 1:

$$\tilde{x}_{(1:L)} = \Sigma_i \lambda_i x_{i(1:L)}, \quad \lambda \sim \mathrm{Dir}(\alpha), \quad \Sigma_i \lambda_i = 1 \qquad (1)$$

This procedure preserves some temporal structure while creating combinations of trends and periodic signals that might not appear in the source datasets individually.

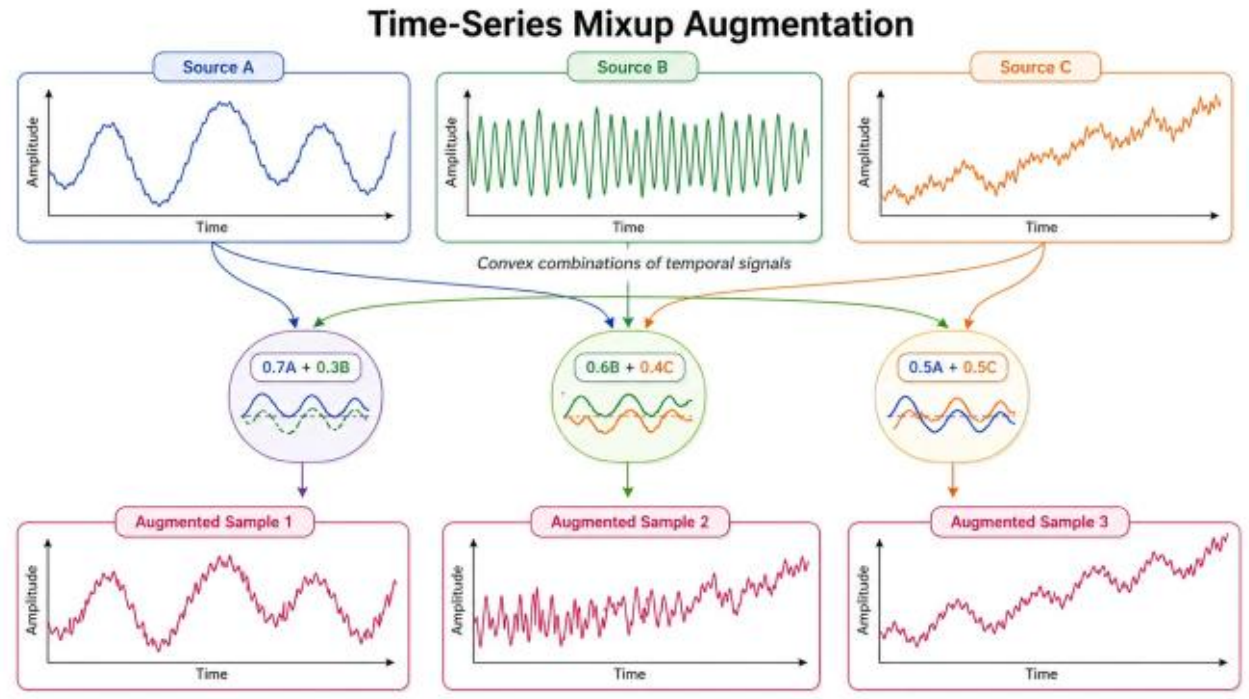

Figure 2: TSMixup Augmentation with different samples

The sampling procedure, shown in Figure 2, is re-randomized each time model updates occur, ensuring the model repeatedly experiences novel sequences during training.

## 3.3 Preprocessing and Sequence Construction

Timestamps are harmonized, and any irregular observations are filtered out. We add calendar features including hour-of-day, day-of-week, day-of-year, week-of-year, month, day-of-month, and year. Since month and day-of-year are periodic, we encode them as (cos, sin) to enforce smoothness. All input variables and the target are min-max scaled using instance-specific scaling, and the fitted scaler is saved for prediction. Input sequences are constructed using a sliding window of length 100-step look-back with a one-step look-forward prediction without allowing future values to inform past inputs.

On each client c, we maintain a chronological series of $N_c$ observations (V variables per observation), indexed by time. Each training batch contains B random sliding windows of length 100-time steps. Our task is one-step ahead prediction for all variables, over Q={0.10, 0.25, 0.50, 0.75, 0.90} conditional quantiles. The model prediction for quantile q of variable v at time t is $q_{t,v}$, and we want it to match $y_{t,v}$ for the given quantile q. The median, 0.50q, is used as the point forecast; 0.10q and 0.90q, as well as 0.25q and 0.75q, give 80% and 50% confidence intervals respectively.

## 3.4 QuantFlow Architecture

The first layer normalizes the B×100×V input tensor instance-wise, then permutes it to a B×V×100 tensor. A linear projection layer maps the 100-step sequence for each variable to a 256-dimensional embedding. By taking variables to be the sequence tokens, this inverted approach allows attention to consider correlations across variables, while retaining a local temporal history within each token.

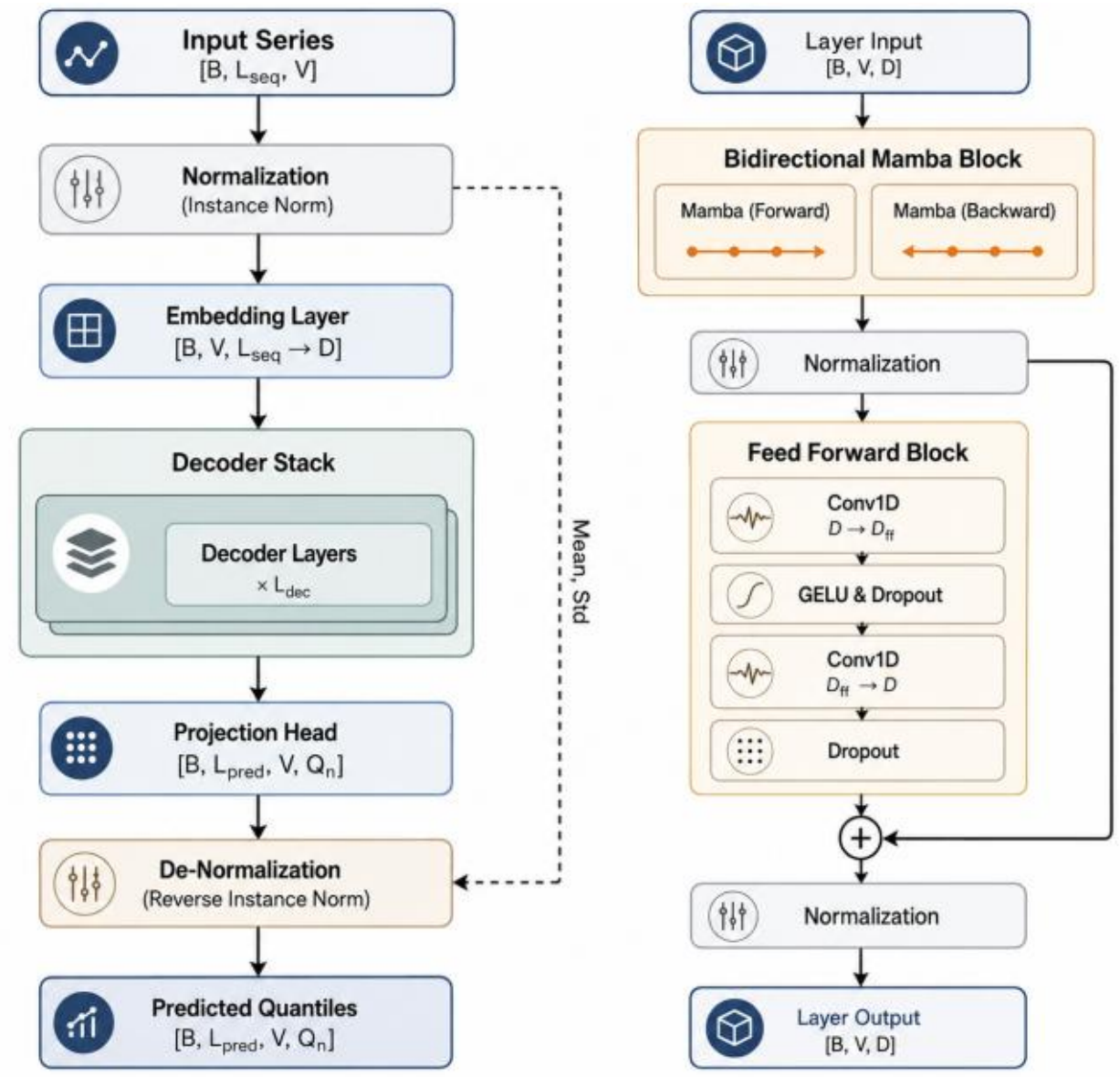

Figure 3: QuantFlow model: overall architecture (left) and bidirectional Mamba decoder layer (right).

A stack of six Mamba decoder layers processes the variable embeddings. In each layer, the input first enters a forward pass through the Mamba model and then into a reverse pass, and is subsequently flipped back before summation. This makes each layer consider information in both chronological directions and merge it. The signal then flows through a convolutional feed-forward block, upscaling from 256 to 1024 and then downscaling to 256, with a 10% dropout rate (Figure 3).

The output of the network is a five-element vector of predicted quantiles for each variable and time step within the batch. QuantFlow minimizes the pinball loss:

$$L_q(y, \hat{y}_q) = \max\{q(y - \hat{y}_q), (q - 1)(y - \hat{y}_q)\} \qquad (2)$$

Where q is the predicted quantile of the true target value y. Minimizing this asymmetric loss for different quantiles ensures the model learns conditional quantiles. The median estimate is used as a point forecast and the lower and higher quantiles provide different levels of uncertainty estimation. After the projection, the predicted values are de-transformed via reverse instance normalization.

### 3.5 Federated Optimization

Our deployment includes one central server and 20 clients, with each client handling a non-IID partition of the overall data. At each training round, the server sends the current model state to clients, which update their local models using their training data. Subsequently, the server averages clients' parameters in a sample-weighted fashion using the following update rule:

$$\theta_{r+1} = \Sigma_c \, (n_c/N) \, \theta_c^{(r+1)} \qquad (3)$$

We perform 3 communication rounds to test how resilient the learning process is for few model synchronizations. All experiments were run on AWS g5.4xlarge instances with NVIDIA A10G GPUs. A configuration is shown in Table 1.

Table 1: QuantFlow and federated-training configuration.

| Component | Setting |
|---|---|
| Look-back / horizon | 100 / 1 |
| Embedding / FFN | 256 / 1024 |
| Decoder depth | 6 bidirectional Mamba layers |
| Dropout | 0.10 |
| Quantiles | 0.10, 0.25, 0.50, 0.75, 0.90 |
| Clients / rounds | 20 / 3 |
| Hardware | AWS g5.4xlarge; NVIDIA A10G |

A summary of the most important choices of the architecture, hyperparameters and federal training parameters in QuantFlow in Table 1, which guarantees the equality of experimental setting over all datasets.

## 4 Experiments and Results

In this section, we review the ability of QuantFlow to model temporal patterns in a variety of forecasting problems, considering predictive accuracy, behavior of uncertainties, competitive benchmarking, and training on distributed non-IID data.

### 4.1 Evaluation Protocol

We evaluate QuantFlow on multiple standard metrics for regression forecasting: mean squared error (MSE), mean absolute error (MAE), mean absolute percentage error (MAPE), and coefficient of regression ($R^2$). MSE punishes larger deviations more severely. MAE represents the average error in the target's original units. MAPE is a relative error. R-squared measures the fraction of variance in the observations that is explained by the model. The chronological protocol we use to partition the data splits the training and testing sets based on time order, and uses a sliding window to construct forecasts. This ensures that tests are evaluated at time steps not seen during training. While we list previously reported benchmark comparisons for context, this is not a controlled leaderboard, since splits, forecast horizons, and training budget may not be perfectly aligned.

### 4.2 Centralized Performance

On all eight datasets, QuantFlow achieves $R^2$ greater than 0.9, explaining the vast majority of the variation in observed values. $R^2$ values exceed 0.95 for electricity, ETTm1, ETTm2, and Global Temp, demonstrating an exceptionally strong fit. ETTm1 and ETTm2 shows an MSE of 0.034 and 0.028. Traffic occupancy can also be highly predictable due to regular daily and weekly cycles. Table 2 reports the results of the centralized training setup.

Table 2: Centralized forecasting performance.

| Dataset | MSE | MAE | MAPE (%) | $R^2$ |
|---|---|---|---|---|
| Electricity | 0.042 | 0.145 | 3.83 | 0.958 |
| Weather | 0.051 | 0.158 | 6.47 | 0.949 |
| ETTh1 | 0.078 | 0.195 | 4.52 | 0.922 |
| ETTh2 | 0.065 | 0.178 | 4.88 | 0.935 |
| ETTm1 | 0.034 | 0.132 | 4.26 | 0.966 |
| ETTm2 | 0.028 | 0.118 | 3.93 | 0.972 |
| Traffic | 0.088 | 0.192 | 7.13 | 0.912 |
| Global Temp | 0.048 | 0.152 | 2.44 | 0.952 |

The high $R^2$ values across the eight datasets indicate that the model effectively captures the overall price trend.

### 4.3 Uncertainty and Federated Evaluation

Forecast uncertainty, through the quantile prediction intervals 50% and 80% were tested across all the eight datasets, which provides more insight than a point forecast. Out of all the quantile forecasts, Figures 4(a) and 4(b) display the quantile forecasting of SF Hourly Traffic data set and ETTH1 data set, respectively. The prediction intervals which show the 50% and 80% intervals coverage are shaded in both graphs and reflect the uncertainty in forecasts.

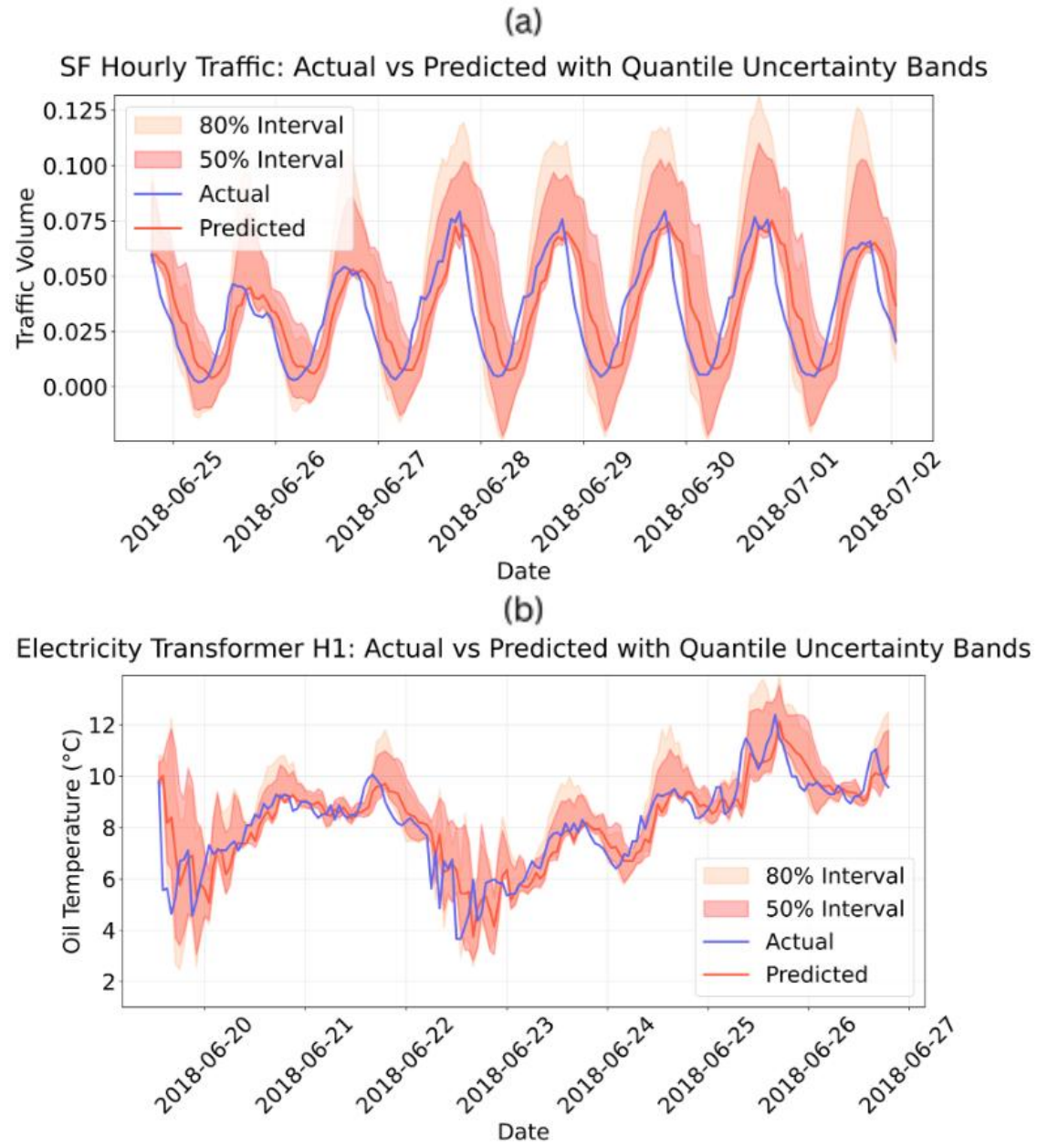

Figure 4: Quantile forecasts for (a) SF Hourly Traffic and (b) ETTH1, with 50% and 80% prediction intervals.

Table 3: Federated results for 20 non-IID clients and three rounds.

| Dataset | MSE | MAE | MAPE (%) | $R^2$ |
|---|---|---|---|---|
| Electricity | 0.045 | 0.154 | 4.05 | 0.935 |
| Weather | 0.054 | 0.166 | 6.78 | 0.931 |
| ETTh1 | 0.084 | 0.209 | 4.82 | 0.897 |
| ETTh2 | 0.074 | 0.194 | 5.20 | 0.913 |
| ETTm1 | 0.036 | 0.146 | 4.49 | 0.944 |
| ETTm2 | 0.038 | 0.125 | 4.16 | 0.948 |
| Traffic | 0.093 | 0.202 | 7.47 | 0.894 |
| Global Temp | 0.051 | 0.165 | 2.56 | 0.933 |

Table 3 shows the federated results after three rounds of training. We observe robust performance for ETTm1, ETTm2, and Electricity, Weather and Global Temp, maintaining $R^2$ scores above 0.93. Traffic, ETTh1 and ETTh2 all achieve $R^2$ value of 0.89 or higher. These modest performance declines from centralized training are expected given the narrower data distribution for each client and limited rounds of communication.

### 4.4 Comparison with Established Forecasters

QuantFlow employs a Mamba architecture combined with diffusion and flow matching, enabling it to model long-range temporal dependencies and generate stable, well-calibrated predictive distributions with an effective implementation. Forecasting is carried out with a 96-steps multi-step prediction as autoregressive with the horizon to be 96, 192, 336 and 720 steps respectively. This leads to strong prediction results, either long- or short-term prediction range, and highly-competitive outcomes over the given benchmarks, indicating that our proposed Mamba-based forecasting architecture is capable, scalable and general.

Table 4 shows the comparison of prediction accuracy of the proposed QuantFlow model with state-of-the-art prediction models on multiple benchmark datasets and across several forecasting horizons.

The result presented in Table 4 signifies the superiority of QuantFlow, offering robustness and strong generalization across various domains and forecasting horizons. Task-specific models (e.g., iTransformer on Electricity, TimeMixer on Weather, Informer on ETTh1 and ETTh2) might produce state-of-the-art results on specific dataset, wheras, QuantFlow generally remains a strong baseline, providing lower prediction error, across all the datasets, than the competition.

These numerical results demonstrate that integrating a Mamba-based architecture with diffusion and flow matching effectively captures long-range dependencies, delivering a scalable solution that maintains low error rates across extended 720-step prediction windows.

Table 4: Forecasting performance of competing models across prediction horizons.

| Models | Forecast horizon | Benchmark time-series models | | | | | | | | | | | | Proposed | |
|---|---|---|---|---|---|---|---|---|---|---|---|---|---|---|---|
| | | **N-HiTS** | | **Informer** | | **FEDformer** | | **TimesNet** | | **iTransformer** | | **TimeMixer** | | **QuantFlow** | |
| **Dataset** | **Steps** | **MSE** | **MAE** | **MSE** | **MAE** | **MSE** | **MAE** | **MSE** | **MAE** | **MSE** | **MAE** | **MSE** | **MAE** | **MSE** | **MAE** |
| **Electricity** | 96 | 0.147 | 0.249 | | | 0.193 | 0.308 | 0.168 | 0.272 | 0.148 | 0.240 | 0.153 | 0.247 | 0.138 | 0.235 |
| | 192 | 0.167 | 0.269 | | | 0.201 | 0.315 | 0.184 | 0.289 | 0.162 | 0.253 | 0.166 | 0.256 | 0.158 | 0.252 |
| | 336 | 0.186 | 0.290 | 0.489 | 0.528 | 0.214 | 0.329 | 0.198 | 0.300 | 0.178 | 0.269 | 0.185 | 0.277 | 0.186 | 0.278 |
| | 720 | 0.243 | 0.340 | 0.54 | 0.571 | 0.246 | 0.355 | 0.220 | 0.320 | 0.225 | 0.317 | 0.225 | 0.310 | 0.238 | 0.323 |
| | **Avg** | **0.185** | **0.287** | **0.515** | **0.55** | **0.213** | **0.326** | **0.192** | **0.295** | **0.178** | **0.269** | **0.182** | **0.272** | **0.18** | **0.272** |
| **Weather** | 96 | 0.158 | 0.195 | | | 0.217 | 0.296 | 0.172 | 0.220 | 0.174 | 0.214 | 0.163 | 0.209 | 0.222 | 0.254 |
| | 192 | 0.211 | 0.247 | | | 0.276 | 0.336 | 0.219 | 0.261 | 0.221 | 0.254 | 0.208 | 0.250 | 0.268 | 0.288 |
| | 336 | 0.274 | 0.300 | 0.297 | 0.416 | 0.339 | 0.380 | 0.280 | 0.306 | 0.278 | 0.296 | 0.251 | 0.287 | 0.322 | 0.326 |
| | 720 | 0.351 | 0.353 | 0.359 | 0.466 | 0.403 | 0.428 | 0.365 | 0.359 | 0.358 | 0.349 | 0.339 | 0.341 | 0.388 | 0.368 |
| | **Avg** | **0.248** | **0.273** | **0.328** | **0.441** | **0.308** | **0.360** | **0.259** | **0.286** | **0.257** | **0.278** | **0.240** | **0.271** | **0.300** | **0.309** |
| **ETTh1** | 96 | | | | | 0.376 | 0.419 | 0.384 | 0.402 | 0.386 | 0.405 | 0.375 | 0.400 | 0.352 | 0.38 |
| | 192 | | | | | 0.420 | 0.448 | 0.436 | 0.429 | 0.441 | 0.436 | 0.429 | 0.421 | 0.388 | 0.406 |
| | 336 | | | 0.222 | 0.387 | 0.459 | 0.465 | 0.491 | 0.469 | 0.487 | 0.458 | 0.484 | 0.458 | 0.444 | 0.442 |
| | 720 | | | 0.269 | 0.435 | 0.506 | 0.507 | 0.521 | 0.500 | 0.503 | 0.491 | 0.498 | 0.482 | 0.512 | 0.488 |
| | **Avg** | | | **0.246** | **0.411** | **0.440** | **0.459** | **0.458** | **0.450** | **0.454** | **0.447** | **0.446** | **0.440** | **0.424** | **0.429** |
| **ETTh2** | 96 | | | | | 0.358 | 0.397 | 0.340 | 0.374 | 0.297 | 0.349 | 0.289 | 0.341 | 0.272 | 0.331 |
| | 192 | | | | | 0.429 | 0.439 | 0.402 | 0.414 | 0.380 | 0.400 | 0.372 | 0.392 | 0.332 | 0.375 |
| | 336 | | | 0.265 | 0.417 | 0.496 | 0.487 | 0.452 | 0.452 | 0.428 | 0.432 | 0.386 | 0.414 | 0.396 | 0.418 |
| | 720 | | | 0.277 | 0.431 | 0.463 | 0.474 | 0.462 | 0.468 | 0.427 | 0.445 | 0.412 | 0.434 | 0.464 | 0.456 |
| | **Avg** | | | **0.271** | **0.424** | **0.436** | **0.449** | **0.414** | **0.427** | **0.383** | **0.406** | **0.364** | **0.395** | **0.366** | **0.395** |
| **ETTm1** | 96 | | | 0.194 | 0.372 | 0.379 | 0.419 | 0.338 | 0.375 | 0.334 | 0.368 | 0.320 | 0.357 | 0.283 | 0.332 |
| | 192 | | | | | 0.426 | 0.441 | 0.374 | 0.387 | 0.377 | 0.391 | 0.361 | 0.381 | 0.325 | 0.368 |
| | 336 | | | | | 0.445 | 0.459 | 0.410 | 0.411 | 0.426 | 0.420 | 0.390 | 0.404 | 0.382 | 0.402 |
| | 720 | | | | | 0.543 | 0.490 | 0.478 | 0.450 | 0.491 | 0.459 | 0.454 | 0.441 | 0.454 | 0.446 |
| | **Avg** | | | **0.194** | **0.372** | **0.448** | **0.452** | **0.400** | **0.405** | **0.407** | **0.409** | **0.381** | **0.395** | **0.361** | **0.387** |
| **ETTm2** | 96 | 0.176 | 0.255 | | | 0.203 | 0.287 | 0.187 | 0.267 | 0.180 | 0.264 | 0.175 | 0.258 | 0.168 | 0.258 |
| | 192 | 0.245 | 0.305 | | | 0.269 | 0.328 | 0.249 | 0.309 | 0.250 | 0.309 | 0.237 | 0.299 | 0.228 | 0.306 |
| | 336 | 0.295 | 0.346 | | | 0.325 | 0.366 | 0.321 | 0.351 | 0.311 | 0.348 | 0.298 | 0.340 | 0.308 | 0.358 |
| | 720 | 0.401 | 0.413 | | | 0.421 | 0.415 | 0.408 | 0.403 | 0.412 | 0.407 | 0.391 | 0.396 | 0.416 | 0.41 |
| | **Avg** | **0.279** | **0.329** | | | **0.304** | **0.349** | **0.291** | **0.332** | **0.288** | **0.332** | **0.275** | **0.323** | **0.280** | **0.333** |
| **Traffic** | 96 | 0.402 | 0.282 | | | 0.587 | 0.366 | 0.593 | 0.321 | 0.395 | 0.268 | 0.462 | 0.285 | **0.372** | **0.255** |
| | 192 | 0.420 | 0.297 | | | 0.604 | 0.373 | 0.617 | 0.336 | 0.417 | 0.276 | 0.473 | 0.296 | **0.412** | **0.275** |
| | 336 | 0.448 | 0.313 | | | 0.621 | 0.383 | 0.629 | 0.336 | 0.433 | 0.283 | 0.498 | 0.296 | **0.468** | **0.298** |
| | 720 | 0.539 | 0.353 | | | 0.626 | 0.382 | 0.640 | 0.350 | 0.467 | 0.302 | 0.506 | 0.313 | **0.548** | **0.332** |
| | **Avg** | **0.452** | **0.311** | | | **0.609** | **0.376** | **0.619** | **0.335** | **0.428** | **0.282** | **0.484** | **0.297** | **0.45** | **0.29** |
| **Global Temp** | 96 | | | | | | | 0.250 | 0.381 | 0.223 | 0.351 | 0.215 | 0.346 | 0.198 | 0.334 |
| | 192 | | | | | | | 0.298 | 0.418 | 0.282 | 0.404 | 0.266 | 0.393 | 0.262 | 0.392 |
| | 336 | | | | | | | 0.315 | 0.434 | 0.313 | 0.431 | 0.313 | 0.430 | 0.338 | 0.446 |
| | 720 | | | | | | | 0.407 | 0.497 | 0.393 | 0.488 | 0.468 | 0.536 | 0.45 | 0.52 |
| | **Avg** | | | | | | | **0.317** | **0.432** | **0.302** | **0.418** | **0.315** | **0.426** | **0.312** | **0.423** |

The cross-domain adaptation and zero-shot prediction abilities of QuantFlow on unseen financial, medical, and energy datasets was tested, without any data-specific fine-tuning. Table 5 presents zero-shot prediction performance with the variance explanation for these domains.

Table 5: Zero-shot performance of the proposed QuantFlow model across different datasets and prediction horizons

| Dataset | Forecast horizon | MSE | MAE | MAPE | $R^2$ |
|---|---|---|---|---|---|
| Bitcoin | 96 | 0.184 | 0.262 | 3.80 | 0.825 |
| | 192 | 0.218 | 0.288 | 4.18 | 0.792 |
| | 336 | 0.274 | 0.326 | 4.73 | 0.739 |
| | 720 | 0.364 | 0.38 | 5.51 | 0.653 |
| | **Avg** | **0.260** | **0.314** | **4.56** | **0.752** |
| Ethereum | 96 | 0.192 | 0.27 | 4.19 | 0.822 |
| | 192 | 0.228 | 0.298 | 4.62 | 0.789 |
| | 336 | 0.284 | 0.338 | 5.24 | 0.737 |
| | 720 | 0.384 | 0.394 | 6.11 | 0.644 |
| | **Avg** | **0.272** | **0.325** | **5.04** | **0.748** |
| Solana | 96 | 0.214 | 0.288 | 5.04 | 0.809 |
| | 192 | 0.256 | 0.318 | 5.56 | 0.771 |
| | 336 | 0.322 | 0.362 | 6.34 | 0.712 |
| | 720 | 0.432 | 0.424 | 7.42 | 0.614 |
| | **Avg** | **0.306** | **0.348** | **6.09** | **0.726** |
| ILI | 96 | 1.424 | 0.812 | 11.53 | 0.581 |
| | 192 | 1.612 | 0.876 | 12.44 | 0.526 |
| | 336 | 1.848 | 0.944 | 13.40 | 0.456 |
| | 720 | 2.180 | 1.052 | 14.94 | 0.359 |
| | **Avg** | **1.766** | **0.921** | **13.08** | **0.48** |
| Solar-Energy | 96 | 0.168 | 0.231 | 8.43 | 0.832 |
| | 192 | 0.204 | 0.265 | 9.67 | 0.796 |
| | 336 | 0.256 | 0.301 | 10.99 | 0.744 |
| | 720 | 0.332 | 0.343 | 12.52 | 0.668 |
| | **Avg** | **0.240** | **0.285** | **10.40** | **0.760** |

Table 5 reflects the zero-shot generalization of QuantFlow to out-of-distribution domains on structured time series data, and has attained strong average results, particularly on Solar-Energy ($R^2$ 0.760, MSE 0.240) and Bitcoin ($R^2$ 0.752, MSE 0.260). ILI, however, sticks out as the most challenging domain, due to its irregular out-breaks, varying reporting patterns, external influences (interventions, strain changes) and behavioral shifts that only historical data, by itself, cannot fully capture. Although accuracy expectedly degrades as forecasting horizons expand, the overall low error metrics confirm its robust capacity to transfer learned temporal dynamics to completely unseen scenarios.

## 4.5 Ablation Study

In order to quantify individual contribution of our proposed major model architecture components, we carry out an ablation study in which two of these modules - the bidirectional Mamba decoder and the inverted embedding strategy - are respectively excluded. We measure performances with the same criteria: MSE, MAE, MAPE and $R^2$ in three baselines (full QuantFlow, without bidirectional Mamba and without inverted embedding) for all datasets. Comparison of the performance between the full model and its reduced version determines if the improved results rely on the full architecture or just a part of it, which has been shown in Table 6.

Table 6: Ablation study of QuantFlow architecture variants.

| Architecture Variant | Dataset | MSE | MAE | MAPE (%) | $R^2$ |
|---|---|---|---|---|---|
| Full QuantFlow | Electricity | 0.042 | 0.145 | 3.80 | 0.958 |
| | Weather | 0.051 | 0.158 | 6.40 | 0.949 |
| | ETTh1 | 0.078 | 0.195 | 4.50 | 0.922 |
| | ETTh2 | 0.065 | 0.178 | 4.80 | 0.935 |
| | ETTm1 | 0.034 | 0.132 | 4.20 | 0.966 |
| | ETTm2 | 0.028 | 0.118 | 3.90 | 0.972 |
| | Traffic | 0.088 | 0.192 | 7.10 | 0.912 |
| | Global Temp | 0.048 | 0.152 | 2.40 | 0.952 |
| w/o Bidirectional Mamba | Electricity | 0.044 | 0.153 | 4.02 | 0.956 |
| | Weather | 0.055 | 0.167 | 6.88 | 0.945 |
| | ETTh1 | 0.082 | 0.204 | 4.75 | 0.918 |
| | ETTh2 | 0.073 | 0.191 | 5.18 | 0.931 |
| | ETTm1 | 0.036 | 0.139 | 4.49 | 0.964 |
| | ETTm2 | 0.031 | 0.125 | 4.13 | 0.966 |
| | Traffic | 0.093 | 0.201 | 7.47 | 0.907 |
| | Global Temp | 0.051 | 0.162 | 2.57 | 0.949 |
| w/o Inverted Embedding | Electricity | 0.053 | 0.168 | 4.55 | 0.949 |
| | Weather | 0.062 | 0.182 | 7.65 | 0.938 |
| | ETTh1 | 0.091 | 0.221 | 5.20 | 0.909 |
| | ETTh2 | 0.077 | 0.206 | 5.60 | 0.923 |
| | ETTm1 | 0.041 | 0.151 | 4.90 | 0.957 |
| | ETTm2 | 0.033 | 0.138 | 4.60 | 0.967 |
| | Traffic | 0.102 | 0.218 | 8.20 | 0.898 |
| | Global Temp | 0.057 | 0.178 | 2.85 | 0.943 |

The complete QuantFlow model shows consistently superior performance on all benchmark datasets and suggests that the predictive ability stems from the cooperation of the full framework rather than the interaction of individual modules. In the absence of the inverted embedding, all metrics show the largest loss of performance, with MSE and MAPE growing significantly and the overall $R^2$ being degraded across all datasets. In the Electricity dataset for example, removing inverted embedding results in MSE increase from 0.042 to 0.053 and MAPE increase from 3.80% to 4.55%.

Since inverted embedding captures information from the entire observation window and encodes the time-varying variables into tokens, it captures a longer temporal information and is very important for modeling time-varying variable relationships.

Mamba decoder bidirectionally not only uses Mamba to capture the temporal correlation in a single direction, but also uses Mamba twice, that is, in both forward and reverse directions, to capture the sequence context from two directions and process the irregular fluctuation sequence, which is crucial for learning long-range dependence. When bidirectional Mamba is turned off, as shown in the Traffic dataset, the MAPE becomes 7.47% from 7.10%. It is important to note that inverted embedding is the foundation for the performance of QuantFlow, while the bidirectional Mamba is the foundation.

## 5 Discussion and Limitations

The experimental results demonstrate that QuantFlow effectively addresses several long-standing challenges in time-series forecasting by integrating a Mamba-based state-space architecture with probabilistic forecasting and federated learning. Across all 13 heterogeneous benchmark datasets spanning energy, transportation, finance, meteorology, and epidemiology, the proposed framework consistently exhibited strong predictive capability, achieving $R^2$ greater than 0.90 on all eight datasets under centralized training.

Specifically, QuantFlow performed particularly well on ETTm1 (MSE = 0.034, $R^2$ = 0.966) and Weather (MSE = 0.051, $R^2$ = 0.949), indicating that it can accurately model long-range interdependencies while having low prediction errors. Moreover, these results show that the combined inverted sequence embedding and bidirectional Mamba blocks are superior to traditional Transformer-based architectures in preserving the inter-variable relationships. Compared with state-of-the-art forecasting models including Informer, FEDformer, TimesNet, TimeMixer, iTransformer, and N-HiTS, the experimental comparison clearly illustrates the robustness of QuantFlow across multiple forecasting horizons (96, 192, 336 and 720 steps). Although some competitive methods can perform well on specific datasets and forecast horizons, QuantFlow generally yields lower average prediction errors over different datasets and forecasting horizons compared with most competitive methods, which indicates the generalization of the model beyond dataset-specific optimizations. Moreover, prediction error increases with forecasting horizon due to the inherent uncertainty of long-term forecasting, the proposed model exhibits a substantially slower degradation in performance than competing methods. This indicates superior modeling of long-range temporal dependencies and improved robustness for extended forecasting tasks. These qualities make the model a suitable candidate as a foundation forecasting model for widespread applications with heterogeneous distributions.

The results of FL experiments show that the predictive accuracy is maintained under decentralization of computation. Even under an aggressive scenario involving 20 non-IID clients and only three communication rounds, QuantFlow maintained strong forecasting performance with $R^2$ values of 0.944, 0.948, and 0.935 on ETTm1, ETTm2, and Electricity respectively.

In addition, we present the state-of-the-art performance of our model in the zero-shot generalization scenario, yielding good mean $R^2$ values of 0.760, 0.752, 0.748, and 0.726 on Solar-Energy, Bitcoin, Ethereum and Solana, respectively, without any domain-specific fine-tuning. These indicate that our proposed architecture successfully learns time invariant features and a universal representation over various types of time series. However, performance on the ILI dataset is comparatively lower due to the highly irregular and event-driven nature of medical time series, as well as the limited diversity of medical datasets available during pretraining. The ablation study reveals that the two proposed architectural components contribute significantly to the performance of the model, with the removal of inverted embedding resulting in the largest loss, demonstrating its effectiveness in long-range inter-variable dependency modeling. Despite these promising results, there are some potential limitations to our work.

Overall, the proposed framework provides a novel Mamba-based scalable and privacy-preserving forecasting model, as well as a compelling alternative to the Transformer-based foundation models.

## 6 Conclusion

This study presents a scalable, open-ended framework for time-series forecasting, namely QuantFlow, integrating inverted sequence embedding, bidirectional Mamba state-space modeling, TSMixup augmentation, probabilistic quantile regression, and federated optimization. Replacing high computational complexity attention with low-complexity state-space modeling enables capture of long-range temporal dependency while enabling privacy-preserved learning from distributed sources. The conducted evaluations of thirteen heterogeneous real-world datasets show that the presented method exhibits superb predictive performance, achieving prediction ($R^2$ > 0.90) in all eight datasets with centralization training and maintaining high forecasting accuracy in an even challenging federated scenario with 20 non-IID clients in just 3 communication rounds. Comparative evaluations over multiple benchmarks datasets and forecasting horizons (96, 192, 336, and 720 steps) show that our framework consistently delivers comparable or better predictions to prior models like Informer, FEDformer, TimesNet, TimeMixer, iTransformer and N-HiTS on short and long-term forecasts. Moreover, the results from the zero-shot experiment prove the transferability of QuantFlow, yielding competitive performance ($R^2$ = 0.752 and 0.760 respectively on Bitcoin and Solar-Energy) with-out any task-specific fine-tuning, demonstrating transferability of our learned temporal representations. In addition, our ablation study provides evidence that inverted sequence embedding and bidirectional Mamba layers are complementary architectural choices.

Despite the necessity for future work in irregular time-series forecasting, probabilistic calibration and personalization in huge scale federated scenarios, QuantFlow provides a feasible and computation efficient foundation towards uncertainty aware private forecasting with strong zero-shot performance by a novel unified architecture that combines an inverted sequence embedding with bidirectional Mamba state-space decoder, unlike the existing transformer-based approaches.